\documentclass[9pt,conference]{IEEEtran}
\IEEEoverridecommandlockouts
\usepackage{cite}
\usepackage{amsmath,amssymb,amsfonts}
\usepackage{algorithmic}
\usepackage{graphicx}
\usepackage{etoolbox,siunitx}
\robustify\bfseries
\usepackage{booktabs}
\usepackage{hyperref}
\usepackage{balance}

\usepackage{textcomp}
\usepackage{xcolor}
\def\BibTeX{{\rm B\kern-.05em{\sc i\kern-.025em b}\kern-.08em
    T\kern-.1667em\lower.7ex\hbox{E}\kern-.125emX}}
\begin{document}

\title{O-EENC-SD: Efficient Online End-to-End Neural Clustering for Speaker Diarization \\
\thanks{This work was supported by the Audible project, funded by French
BPI and partly supported by ANR Project SAROUMANE (ANR-22-CE23-
0011). Also, it was performed using HPC resources from GENCI-IDRIS.}
}

\author{
    \IEEEauthorblockN{\textit{
    Elio Gruttadauria\IEEEauthorrefmark{1}, 
    Mathieu Fontaine\IEEEauthorrefmark{1},
    Jonathan Le Roux\IEEEauthorrefmark{2}, 
    Slim Essid\IEEEauthorrefmark{1}
    }
}
\vspace{0.5cm}
\IEEEauthorblockA{
    \IEEEauthorrefmark{1}LTCI, T\'el\'ecom Paris, Institut polytechnique de Paris, Palaiseau, France\ 
    \\
    \IEEEauthorrefmark{2}Mitsubishi Electric Research Laboratories (MERL), Cambridge, MA, USA\
    }
}


\newcommand{\se}[1]{{\textcolor{green}{#1}}}
\newcommand{\secor}[2]{{\textcolor{orange}{[#1 $\rightarrow$ #2]}}}
\newcommand{\secmt}[1]{{\textcolor{blue}{[SE: #1]}}}

\maketitle

\begin{abstract}
We introduce O-EENC-SD: an end-to-end online speaker diarization system based on EEND-EDA, featuring a novel RNN-based stitching mechanism for online prediction. In particular, we develop a novel centroid refinement decoder whose usefulness is assessed through a rigorous ablation study.
Our system provides key advantages over existing methods: a hyperparameter-free solution compared to unsupervised clustering approaches, and a more efficient alternative to current online end-to-end methods, which are computationally costly.
We demonstrate that O-EENC-SD is competitive with the state of the art in the two-speaker conversational telephone speech domain, as tested on the CallHome dataset.
Our results show that O-EENC-SD provides a great trade-off between DER and complexity, even when working on independent chunks with no overlap, making the system extremely efficient.
\end{abstract}


\section{Introduction}
Speaker diarization (SD) is an important speech processing task which aims at estimating the speech activities of each speaker in a real-world recording~\cite{park_review_2021}.
In online SD, data is processed on-the-fly in a streaming fashion and predictions must be causal.
In contrast, in traditional or offline SD, the whole recording is available as the input of the system.

While SD is traditionally framed as a clustering problem \cite{sahidullah_speed_2019, landini_analysis_2021}, recent developments pushed forward a novel end-to-end paradigm \cite{fujita_end--end_2019, horiguchi_end--end_2020, horiguchi_towards_2021}.
The benefits of the end-to-end neural diarization (EEND) family of models include the direct optimization of the diarization error and better handling of overlapped speech.
Nevertheless, clustering remains a flexible tool both for offline and online inference.
In offline SD, clustering has been used to reduce the computational burden of end-to-end models \cite{kinoshita_integrating_2021, kinoshita_advances_2021} or to solve the empirical limitations on the maximum number of speakers the models can handle \cite{yang_robust_2022}.
Clustering has also been used to adapt models to work in an online fashion \cite{coria_overlap-aware_2021, gruttadauria_online_2024}: rather than the entire sequence being processed altogether, it is split into chunks whose predictions are then stitched back together.
As end-to-end models are commonly trained under a permutation-invariant training (PIT) paradigm \cite{fujita2_end--end_2019}, the order of speaker activities in the output is unknown.
In this context, the role of clustering is to solve the speaker permutation problem in-between blocks.

Another approach to solve the permutation problem is to include an overlap between consecutive blocks, as the shared portion can be used to reorder the speakers.
This shared portion is commonly referred to as the \textit{buffer}.
The drawback of this approach is the repeated computation on the overlap region, which can be reduced with the use of more advanced buffering strategies \cite{xue_online_2021, xue_online_2021-1}.
Buffering strategies encompass different sampling techniques to compose the shared portion between chunks, reaching performances competitive to fully causal models \cite{horiguchi_online_2022}.
Even so, the buffer size considered in state-of-the-art systems tends to be quite large ($>= 100 s$), which entails a substantial computational cost (in terms of operations per second), thereby hindering their applicability to settings requiring efficient realizations, typically when the solution needs to run at the edge. This leaves the need for robust \textbf{lightweight} online variants, especially variants exploiting smaller buffers.

\begin{figure}[t]
    \centering
    \includegraphics[width=1.0\linewidth]{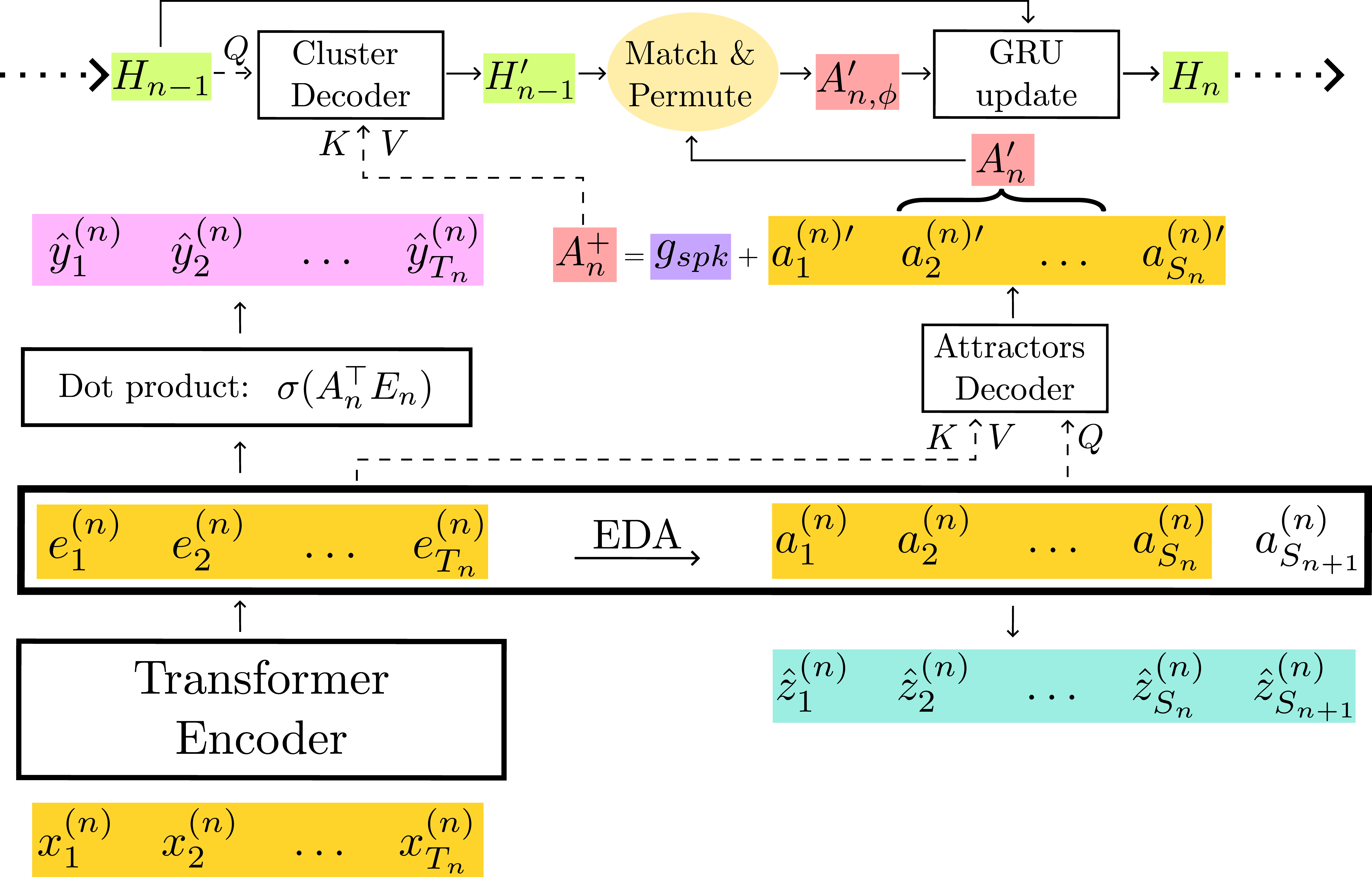}
    \vskip -2mm
    \caption{Outline of O-EENC-SD, detailing the processing of a single chunk.}
    \label{fig:fig1}
    \vskip -3mm
\end{figure}

\noindent\textbf{Contributions:} In this work, we present a novel online speaker diarization system based on end-to-end neural diarization with encoder-decoder attractors (EEND-EDA) \cite{horiguchi_end--end_2020}  and leveraging online neural clustering to solve the speaker permutation problem between predictions of consecutive chunks. The system, illustrated in Fig.~\ref{fig:fig1},
is competitive with the state of the art on online speaker diarization in the two-speaker conversational telephone speech (CTS) domain, as measured by the diarization error rate (DER).
The strength of this work relies on the improved trade-off between performance and computational complexity, as shown by the marginal degradation of the DER when constraining the computational budget.
A particular contribution of this work is the introduction of a new centroid refinement decoder, used to improve the accuracy of the neural clustering.
To show the importance of each component of the model, we conduct an ablation study under different settings of latency and computational budget.
Finally, we analyze the performance of O-EENC-SD when neither data nor computation is shared between subsequent chunks, opening the path to future works focusing on real-time SD on edge devices.
The code to reproduce the results of this work is freely available.\footnote{\url{https://github.com/egruttadauria98/O-EENC-SD}} 

\begin{figure*}[t]
    \centering
    \includegraphics[width=1.0\linewidth]{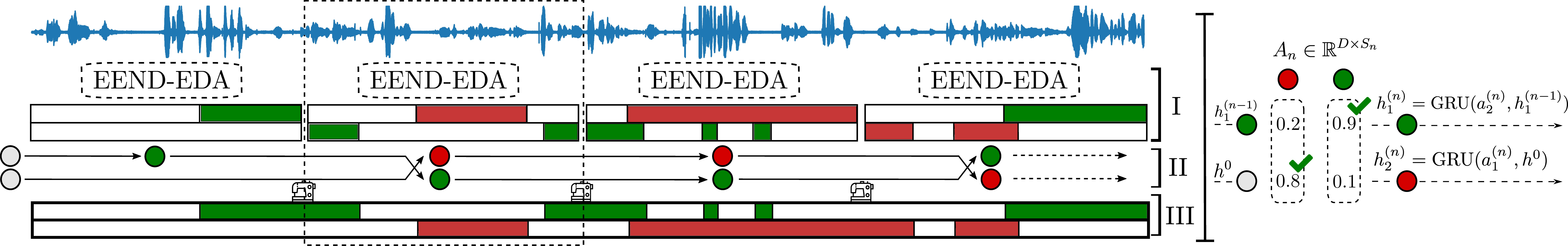}
    \caption{\textbf{Left}: High-level view of the stitching process on non-overlapping chunks. The EEND-EDA model estimates $\hat{Y}_n$ (I) and the relative attractors, matched to the centroids (II). Based on the optimal permutation between centroids and attractors, the local predictions are permuted accordingly and stitched to the previous chunks to form the clustering output (III). \textbf{Right}: Zoom on the matching process for the second chunk (enclosed in dashed box, $n=2$). Two attractors need to be matched: the second is matched to the only centroid (\textit{green}), while the first is matched to $h^0$ (\textit{grey}), creating a new centroid. 
    }
    \label{fig:fig2}
    \vskip -3mm
\end{figure*}

\section{Related Work}
In speaker diarization, few models are online by nature \cite{han_bw-eda-eend_2021, zhang_fully_2019}. 
More commonly, online SD is performed with offline models adapted to handle streaming data \cite{coria_overlap-aware_2021, xue_online_2021, xue_online_2021-1, horiguchi_online_2022, gruttadauria_online_2024}.
To do so, the data is divided into chunks to be processed ``locally",  before combining the output of local (chunk-level) predictions through a stitching process.

In recent years, the end-to-end paradigm has gained a lot of traction in the landscape of offline SD.
Currently, several alternative methods are available to adapt end-to-end diarization (EEND) models to operate online.
Coria et al.~\cite{coria_overlap-aware_2021} proposed using a stitching method based on unsupervised online clustering of the speaker embeddings associated with each predicted speech activity.
The advantage of clustering is to enable handling more speakers in the whole recording than what is allowed by the EEND model in a single chunk.
As an unsupervised method, clustering can be extremely flexible, but also hard to optimize, and a gap between training and inference is usually observed.
Furthermore, clustering usually requires extensive domain-specific hyperparameter tuning to achieve high performance.

To optimize an EEND model to work online in an end-to-end fashion, a speaker-tracing buffer (STB)~\cite{xue_online_2021, xue_online_2021-1} can be used.
In its simplest form, using a first-in-first-out (FIFO) strategy, it is comparable to having an overlap between consecutive chunks with a hop size smaller than the chunk size.
More advanced strategies involve heuristics for selecting past frames to summarize as much information into a limited buffer.
In \cite{horiguchi_online_2022}, using speaker-balanced sampling probabilities with a 100-s long frame-wise STB was shown to even improve the DER on CallHome compared to a fully causal model, from 15\% to 14.93\%.
Unfortunately, using a 100-s buffer with a hop size of 1 s is not feasible for many applications with limited computational resources.
As the buffer needs to preserve the information of all the speakers seen so far in the form of acoustic features to be further processed, the buffer size is critical for performance (see Table 3 of \cite{horiguchi_online_2022}).
In contrast, neural clustering \cite{yang_robust_2022} stores speaker information in a much more compact format: a centroid for each seen speaker, evolving from step to step.


\section{Proposed System}
The proposed system processes the data in chunks, as shown in Figs.~\ref{fig:fig1} and \ref{fig:fig2}.
Online diarization is obtained by sequentially stitching the chunk-predictions together, thereby solving the speaker permutation problem.
New speakers might also be found as data is processed.
Roughly, the system is composed of two main components: an EEND-EDA model \cite{horiguchi_end--end_2020} to obtain local predictions from each chunk, and a recurrent neural network (RNN)~\cite{schmidt_recurrent_2019} that performs online neural clustering for the stitching. 
To refine the latent representation of the speakers, two transformer decoders \cite{vaswani_attention_2023} are used: an \textit{attractor refinement decoder} and a \textit{centroid refinement decoder}.
The latter is one of the novel contributions of this work.

\subsection{EEND-EDA}
The EEND-EDA model processes acoustic features $X \in \mathbb{R}^{D^\prime\times T_{n}}$ into frame embeddings $E_n \in \mathbb{R}^{D\times T_{n}}$, where $n$ is the chunk index and $T_n$ is the number of frames in a chunk.
The EDA component estimates attractors $A_n = [a^{(n)}_1, \dots, a^{(n)}_{S_n}] \in \mathbb{R}^{D\times S_{n}}$ in an autoregressive fashion, where $S_{n}$ is the number of speakers in the chunk.
To let the network determine that there are no more speakers, existence probabilities $\hat{Z}_n = [\hat{z}^{(n)}_1, \dots, \hat{z}^{(n)}_{S_n+1}] \in (0, 1)^{S_{n}+1}$ are computed.
The speaker probabilities are estimated for each frame as the dot product between the frame embeddings and the attractors, followed by a sigmoid activation: $\hat{Y}_n = \sigma(A_{n}^{\top} E_n) \in (0, 1)^{S_{n} \times T_{n}}$, where $\hat{Y}_n = [\hat{y}^{(n)}_1, \dots, \hat{y}^{(n)}_{T_n}]$.
To train the model, two losses are used: an existence loss $L^{\text{chunk}}_{\text{exist}}$ on the attractors
and a diarization loss $L^{\text{chunk}}_{\text{diar}}$. 
Both losses are implemented using the binary cross-entropy (BCE):
\begin{equation}
    L^{\text{chunk}}_{\text{exist}} = \frac{1}{N} \sum_{n=1}^{N} \text{BCE}(\hat{Z}_n, Z_n) , \quad Z_n = [\underbrace{1, \dots, 1}_{S_n}, 0] ,
\end{equation}
\begin{equation}
    L^{\text{chunk}}_{\text{diar}} = \frac{1}{N} \sum_{n=1}^{N} \frac{1}{T_n S_n} \min_{\phi \in \Phi_n} \sum_{t=1}^{T_n} \text{BCE}(\hat{y}^{(n)}_{t}, P_\phi y_{t}^{(n)}),
\end{equation}
\begin{equation}
    L^{\text{chunk}}_{\text{EEND-EDA}}=L^{\text{chunk}}_{\text{diar}}+L^{\text{chunk}}_{\text{exist}},
\end{equation}
where $N$ is the total number of chunks, $\Phi_n$ is the set of all permutations $\phi$ of $S_n$ speakers, and $P_\phi$ denotes the permutation matrix associated with $\phi$.
During training, the EEND-EDA model is also run on the entire audio and the losses are applied globally, similarly to \cite{horiguchi_online_2022}:
\begin{equation}
    L^{\text{global}}_{\text{exist}} = \text{BCE}(\hat{Z}, Z), Z = [\underbrace{1, \dots, 1}_{S}, 0] ,
\end{equation}
\begin{equation}
    L^{\text{global}}_{\text{diar}} = \frac{1}{T S} \min_{\phi \in \Phi} \sum_{\tau=1}^{T} \text{BCE}(\hat{y}_{\tau}, P_{\phi}y_{\tau}),
\end{equation}
\begin{equation}
    L^{\text{global}}_{\text{EEND-EDA}}=L^{\text{global}}_{\text{diar}}+L^{\text{global}}_{\text{exist}},
\end{equation}
where $T$ and $S$ are the number of frames and the number of speakers in the entire recording, respectively.

\subsection{Online neural clustering}
Online diarization is performed by finding the correspondence of attractors across consecutive chunks.
Neural clustering \cite{yang_robust_2022} leverages RNNs to model the evolution of speaker centroids,
each modeled by a different RNN.
The hidden state of the RNN represents the centroid of the cluster.
All RNNs are initialized with a common trainable embedding $h^{0}$.
Rather than using a fixed rule to update the centroids, RNNs can provide a more robust and flexible alternative.
Specifically, we use gated recurrent units (GRUs) \cite{cho_learning_2014}.

The online neural clustering is trained by classifying each attractor of the chunk to one of the known $C$ centroids or to $h^{0}$, meant to represent an average speaker.
Using $h^{0}$ as an option also allows online clustering to be framed as a closed set and fully differentiable problem.
A cross-entropy loss is used to learn the correct assignment of chunk attractors to centroids:


\begin{equation}
    L_{\text{CE}}^{\text{cluster}} = \frac{1}{N} \sum_{n=1}^{N} \Big( \frac{1}{S_n} \sum_{i=1}^{S_n} \sum_{j=1}^{C} -r_{ij}^{(n)} \cdot \log(p_{ij}^{(n)}) \Big) ,
\end{equation}
with $(p_{ij}^{(n)})_{j=1,\dots,C} = \sigma_c \big(H_{n-1}^\top  a_i^{(n)} \big)$, where $\sigma_c$ denotes the softmax over the centroids and $ H_{n-1} = [h^{(n-1)}_1, \dots, h^{(n-1)}_C] \in \mathbb{R}^{D\times C}$ the hidden states of the previous step.
For an attractor $i$ whose speaker identity is represented by centroid $j$, $r_{ij}^{(n)}$ is $1$, otherwise it is $0$.
Another option that we propose to supervise the learning of the neural clustering is to apply $L^{\text{global}}_{\text{diar}}$ to the output of the stitching process, depicted in Fig.~\ref{fig:fig2}.
We will refer to this loss as $L^{\text{cluster}}_{\text{diar}}$.

At the end of each step, the GRUs of the active speakers are updated with the matched attractor using a teacher-forcing strategy~\cite{williams_learning_1989}.
The hidden states corresponding to the GRU of non-active speakers are not updated. Thus,
\begin{equation}
    h_j^{(n)} = 
    \begin{cases} 
    h_j^{(n-1)} & \mathrm{if} \; \forall i, \quad r_{ij}^{(n)} = 0 \\
    \text{GRU}(a_i^{(n)}, h_j^{(n-1)}) & \mathrm{if} \; \exists i, \quad r_{ij}^{(n)} = 1
    \end{cases},
    \label{eq:update}
\end{equation}
where $1 \leq i \leq S_n$ and $1 \leq j \leq C$.

\subsection{Refinement decoders}
In the EEND-EDA model, the frame features are encoded by a stack of transformer layers \cite{vaswani_attention_2023}.
Attractors are then estimated in an autoregressive fashion using an LSTM \cite{hochreiter1997long} encoder-decoder system.
As attractors are optimized to minimize the diarization error within a chunk, \cite{horiguchi_online_2022} suggested using a transformer decoder to refine the attractors for inter-chunk stitching.
Rather than using all past frame embeddings, we use only the ones of the current chunk as input to the decoder to make computation lightweight:
\begin{equation}
    A^{\prime}_n = \text{TransformerDecoder}(A_n, E_n, E_n) \in \mathbb{R}^{D\times S_{n}}.
\end{equation}

A novel contribution of this work is the centroid refinement decoder.
In online clustering, the update of the centroids usually happens after the new data points have been assigned to their corresponding clusters. 
The idea of the centroid refinement decoder is to perform a contextual update given the current attractors to be matched.
To do so, the decoder uses a cross-attention mechanism between the centroids $H_{n-1}$ and the concatenation of a trainable embedding $g_{\text{spk}} \in \mathbb{R}^D$, referred to as \textit{ghost speaker},  with the attractors in the new chunk $A_{n}$:
\begin{equation}
    H^{\prime}_{n-1} = \text{TransformerDecoder}(H_{n-1}, A^+_n, A^+_n) \in \mathbb{R}^{D\times S_n},
\end{equation}
where $A^+_n=\text{Concat}(g_{\text{spk}}, A^{\prime}_n) \in \mathbb{R}^{D\times (S_{n}+1)}$.

The rationale behind the use of a ghost speaker is to add a degree of freedom to the refinement of the centroids of non-active speakers.
In the latent space of the attention mechanism, queries are updated as a convex combination of keys and values.
For active speakers, the decoder can push the centroid towards the correct attractor to facilitate the stitching step.
For non-active speakers, the ghost speaker represents an extra option to attend to rather than the chunk's attractors.


\section{Experimental Setup}

\subsection{Data}
As we focus on conversational telephone speech (CTS), we evaluate O-EENC-SD on the CallHome corpus \cite{przybocki_mark_2000_2001}, using a 0.25-s tolerance collar to compare our results to previous works.\ 
For training, we distinguish between pretraining on simulated data and fine-tuning on real data.
For the simulated data, we rely on the licence-free CallHome and CallFriend corpora available on TalkBank \cite{noauthor_talkbank_nodate} as input data, which are provided in stereo format and with no annotation.
The data simulation pipeline is as follows: the stereo channels are analyzed individually with a pretrained diarization model to filter out the ones with more than one speaker.
The remaining channels are combined randomly into simulated phone conversations.
To improve the realism, speakers are sampled from the same language group.
For the fine-tuning part, we use the development set of the CallHome corpus.

\subsection{Architecture and Training Details}
The configuration of the EEND-EDA model, and of the input features to the model, is the same as described in \cite{horiguchi_end--end_2020}.\
For the pretraining phase of O-EENC-SD, the EEND-EDA model is initialized as the average of the checkpoints of the last 10 epochs after being trained on simulated conversations (option 2), as described in \cite{landini_simulated_2022}.
The neural clustering component is initialized randomly.
The learning rate is set to $10^{-4}$ for the pre-training phase and $10^{-5}$ for the fine-tuning phase.
Before pre-training with the desired buffer size, all models are first trained on 10-s audio segments, split into 10 non-overlapping 1-s chunks.
Our experiments show that this step leads to performance gains in the fine-tuned model.\\
Similarly to \cite{horiguchi_online_2022}, we use both a global and a chunk-level EEND-EDA loss.
In total, four loss terms are used: two for the EEND-EDA model ($L_{\text{EEND-EDA}}$ global and $L_{\text{EEND-EDA}}$ chunk) and two for the neural clustering ($L_{\text{CE}}$ and $L_{\text{diar}}$).
The total loss is computed as:
\begin{equation}
    L=L_{\text{EEND-EDA}}^{\text{global}}+10 L_{\text{EEND-EDA}}^{\text{chunk}}+L_{\text{CE}}^{\text{cluster}}+L_{\text{diar}}^{\text{cluster}} ,\end{equation}
where $10$ is the number of chunks into which the train data is usually split.
During pre-training, we have found beneficial to only use the classification loss for the neural clustering.
Models are trained to work at different latencies by modifying the attention mask of the EEND-EDA encoder.
During inference, we only consider first-in-first-out (FIFO) \cite{xue_online_2021, xue_online_2021-1} as the buffering strategy.

\section{Results}
In this work, we refer to the \textbf{entire} context used to process a chunk (including past observations and innovation) as \textit{buffer}.
As such, the buffer size is always greater than or equal to the latency, equivalent to the online processing unit in \cite{horiguchi_online_2022}.

Table \ref{table:models} collects the results of O-EENC-SD on CallHome to compare our proposal to state-of-the-art systems. 
All shown variants are trained with a 50-s buffer size, except for models in Table \ref{table:efficient}, where the buffer size is equal to the latency.
Unless otherwise specified, all models are trained with the same latency as used at test time.

\subsection{Comparison to the state of the art}
Our system demonstrates competitive performance with previous proposals, particularly when higher latency is allowed.
Using a \mbox{100-s} FIFO buffer, O-EENC-SD achieves a DER of 9.53\% and 9.50\% at 5-s and 10-s latency, respectively.
The best model achieves a DER of 9.33\% at 5-s latency when all past frames are included in the buffer.
Our experiments reveal that O-EENC-SD achieves better performance when trained with higher latency values.
Indeed, when performing inference with a 1-s latency, using a 100-s buffer size, the model trained with 5-s latency achieves 11.96\% DER, better than the 12.47\% of the 1-s latency model.
Even with a significantly smaller 25-s buffer size, the 5-s latency model reaches a better 12.14\% DER when tested at 1-s latency than the 1-s latency model using a 100-s buffer size.
A possible explanation is that higher latency provides a stronger training effect on the EEND-EDA model to make better predictions at the chunk level.
On the contrary, at low latency, the stitching is easier as the overlap portion between chunks is greater, leading to a smaller training effect.
More specifically, the average EEND-EDA performance on a chunk is 11.84\% DER when trained at 1-s latency and 8.89\% at 5-s latency.

Training strategy improvements can lower the DER considerably in EEND-like models.
In the top section of Table \ref{table:models}, it is shown that the performance of EEND-EDA+FW-STB can be improved from 12.70\% DER to 9.08\% with the use of variable chunk-size training (VCT) \cite{horiguchi_online_2022}, paired with improvements in the sampling probabilities of the STB.
The profound effect of VCT stresses how much EEND-EDA performance on small chunks is essential for the overall online DER.
From our experiments, high-latency training appears to be a good strategy to improve chunk performance, but it presents a domain shift with low-latency inference.
We believe future works can bridge this gap and make online neural clustering competitive even at low latency requirements.

\begin{table}[t]
\centering
\fontsize{7}{9}\selectfont 
\sisetup{
detect-weight, 
mode=text, 
tight-spacing=true,
round-mode=places,
round-precision=0,
table-format=3,
table-number-alignment=center
}
\caption{DERs (\%) on CallHome 2-spks test set with 0.25s collar tolerance.\
}
\label{table:models}
\begin{tabular}{lSS[table-format=2]S[round-precision=2,table-format=2.2]}
\toprule
\textbf{Model} & \textbf{Buffer (s)} & \textbf{Latency (s)} & \textbf{DER (\%)} \\
\midrule
BW-EDA-EEND \cite{han_bw-eda-eend_2021} & 10 & 10 & 11.82 \\
EEND-EDA+FW-STB \cite{xue_online_2021-1} & 101 & 1 & 12.70 \\
EEND-EDA+FW-STB\textsuperscript{\dag} \cite{horiguchi_online_2022} & 101 & 1 & 9.08 \\
EEND-GLA-Small+BW-STB \cite{horiguchi_online_2022} & 100 & 1 & \bfseries 9.01 \\
EEND-GLA-Large+BW-STB \cite{horiguchi_online_2022} & 100 & 1 & 9.20 \\
\midrule
\midrule
O-EENC-SD & $\infty$ & 10 & 9.50 \\ 
O-EENC-SD & 100 & 10 & 9.50 \\ 
O-EENC-SD & 10 & 10 &  10.75 \\ 
\midrule
O-EENC-SD & $\infty$ & 5 & 9.33 \\ 
O-EENC-SD & 100 & 5 & 9.53 \\
O-EENC-SD & 5 & 5 & 13.20 \\ 
\midrule
O-EENC-SD$^*$ & $\infty$ & 1 & 11.45 \\ 
O-EENC-SD$^*$ & 100 & 1 & 11.96 \\ 
O-EENC-SD$^*$ & 50 & 1 & 12.54 \\ 
O-EENC-SD$^*$ & 25 & 1 & 12.14 \\ 
O-EENC-SD$^*$ & 10 & 1 & 14.5 \\ 
O-EENC-SD$^*$ & 5 & 1 &  19.99 \\ 
\midrule
O-EENC-SD Base & 100 & 1 & 15.38 \\ 
\quad + Attractors decoder & 100 & 1 & 15.07 \\ 
\quad + Centroid decoder & 100 & 1 & 12.69 \\ 
O-EENC-SD & 100 & 1 & 12.47 \\ 
\bottomrule
\end{tabular}
\vskip 1mm
\parbox{\linewidth}{
\footnotesize \textsuperscript{\dag} Reproduced results with improved training methodology.\\
$^*$ Model trained with a 5-s latency and a 50-s buffer size.
}
\end{table}

\subsection{Architecture Ablation}
The bottom of Table \ref{table:models} also includes an ablation study on the model architecture at 1-s latency.
We find that both decoders do improve performance over the base model.
When the decoders are used jointly, O-EENC-SD exhibits the best results, increasing the clustering accuracy from 90\% to 93\% over the base model.
Also, the performance of the EEND-EDA model in each chunk improves, decreasing the DER from 12.68\% to 11.84\%.
Overall, the online clustering performance improves from 15.38\% to 12.47\% DER.

\begin{table}[t]
\fontsize{6.5}{7}\selectfont 
\centering
\sisetup{
detect-weight, 
mode=text, 
tight-spacing=true,
round-mode=places,
round-precision=2,
table-format=2.2,
table-number-alignment=center
}
\caption{Low-computation results on CallHome 2-spks test set.\\
$L$ denotes the number of context blocks in  \cite{han_bw-eda-eend_2021}.}
\label{table:efficient}
\begin{tabular}{lcS[round-precision=0,table-format=2]SS}
\toprule
\textbf{Model} & \!\!\textbf{Complexity}\!\! & \!\textbf{Buffer (s)}\!  & \!\textbf{DER (\%)}\! & \!\!\textbf{Accuracy (\%)}\!\! \\
\midrule
BW-EDA-EEND, $L\!=\!\infty$ \cite{han_bw-eda-eend_2021} & $\mathcal{O}(T)$ & 10 & 11.82 & {N/A} \\
BW-EDA-EEND, $L\!=\!1$ \cite{han_bw-eda-eend_2021} & $\mathcal{O}(1)$ & 10 & 16.18 & {N/A} \\
\midrule
\midrule
O-EENC-SD Base & $\mathcal{O}(1)$ & 5 & 13.94 & 93.14  \\ 
O-EENC-SD & $\mathcal{O}(1)$ & 5 & 13.20 & 96.16   \\ 
\midrule 
O-EENC-SD Base & $\mathcal{O}(1)$ & 10 & 12.93 & 94.24 \\ 
O-EENC-SD & $\mathcal{O}(1)$ & 10 & \bfseries 10.75 & \bfseries 97.17  \\ 
\bottomrule
\end{tabular}
\vskip 1mm
\parbox{\linewidth}{
\footnotesize Latency is set to the buffer size for all models.\\
Accuracy refers to the ratio of correctly classified attractors to the total.
}
\end{table}

\subsection{Efficiency}
The proposed O-EENC-SD system particularly shines when tested at low buffer sizes and on non-overlapping chunks, making the system efficient from a computational perspective.
By contrast, the paradigm of the previous methods \cite{xue_online_2021, xue_online_2021-1, horiguchi_online_2022} requires a long buffer to work properly.
The only other EEND-based online model that does not use a buffer is BW-EDA-EEND \cite{han_bw-eda-eend_2021}, which implements a version of the EEND-EDA encoder motivated by Transformer-XL \cite{dai_transformer-xl_2019}: for every chunk, the hidden states of all layers are cached after being computed, to be used in the computation of the next $L$ chunks.
If $L=\infty$, all previously cached hidden states are used in the current chunk and the time complexity of the attention mechanism is linear with respect to the sequence length $T$.
If $L=1$, only the hidden states from the previous chunk are used and the time complexity does not depend on the sequence length.
For O-EENC-SD, when the latency is set equal to the buffer size as in Table \ref{table:efficient}, each chunk is considered independently.
As the RNN update does not depend on $T$, the computation for a single chunk has constant time complexity. 
Table~\ref{table:efficient} shows that O-EENC-SD with a 10-s buffer size outperforms BW-EDA-EEND with $L=\infty$ by a significant margin, lowering the DER from 11.82\% to 10.75\%.
Noticeably, all the model versions in Table~\ref{table:efficient} achieve better performance than BW-EDA-EEND with $L=1$.

To conclude, we also want to remark on the impressive clustering accuracy of O-EENC-SD when trained with small buffer sizes, even surpassing 97\% for the 10-s buffer size model.
For models trained with a 50-s buffer size, the average accuracy is instead around 93\%.
As for the high-latency training, it seems that the online neural clustering prefers harder training strategies, which will be leveraged in future work to make the neural clustering paradigm stronger at low-latency requirements.


\section{Conclusions}
In this work, we have proposed O-EENC-SD, a novel online speaker diarization method providing a great trade-off between DER and latency/complexity.
Our results show that online neural clustering is competitive with previous systems, thanks in particular to the proposed centroid refinement decoder and the downstream diarization loss.
Future work will focus on leveraging the strong potential of 
high-latency training for low-latency inference.
To conclude, we believe online neural clustering has the potential to become state-of-the-art in online speaker diarization, combining efficiency and performance into a single paradigm. 

\section*{Acknowledgments}
We would like to thank Chenyu Yang for their implementation of the offline neural clustering method, and Federico Landini for the checkpoints of the pretrained EEND-EDA model.

\balance
\bibliographystyle{IEEEtran}
\bibliography{bib}

\begin{thebibliography}{10}
\providecommand{\url}[1]{#1}
\csname url@samestyle\endcsname
\providecommand{\newblock}{\relax}
\providecommand{\bibinfo}[2]{#2}
\providecommand{\BIBentrySTDinterwordspacing}{\spaceskip=0pt\relax}
\providecommand{\BIBentryALTinterwordstretchfactor}{4}
\providecommand{\BIBentryALTinterwordspacing}{\spaceskip=\fontdimen2\font plus
\BIBentryALTinterwordstretchfactor\fontdimen3\font minus
  \fontdimen4\font\relax}
\providecommand{\BIBforeignlanguage}[2]{{%
\expandafter\ifx\csname l@#1\endcsname\relax
\typeout{** WARNING: IEEEtran.bst: No hyphenation pattern has been}%
\typeout{** loaded for the language `#1'. Using the pattern for}%
\typeout{** the default language instead.}%
\else
\language=\csname l@#1\endcsname
\fi
#2}}
\providecommand{\BIBdecl}{\relax}
\BIBdecl

\bibitem{park_review_2021}
T.~J. Park, N.~Kanda, D.~Dimitriadis, K.~J. Han, S.~Watanabe, and S.~Narayanan,
  ``A {Review} of {Speaker} {Diarization}: {Recent} {Advances} with {Deep}
  {Learning},'' 2021.

\bibitem{sahidullah_speed_2019}
M.~Sahidullah, J.~Patino, S.~Cornell, R.~Yin, S.~Sivasankaran, H.~Bredin,
  P.~Korshunov, A.~Brutti, R.~Serizel, E.~Vincent \emph{et~al.},
  ``\BIBforeignlanguage{en}{The {Speed} {Submission} to {DIHARD} {II}:
  {Contributions} \& {Lessons} {Learned}},'' 2019.

\bibitem{landini_analysis_2021}
F.~Landini, O.~Glembek, P.~Matějka, J.~Rohdin, L.~Burget, M.~Diez, and
  A.~Silnova, ``\BIBforeignlanguage{en}{Analysis of the {BUT} {Diarization}
  {System} for {VoxConverse} {Challenge}},'' 2021.

\bibitem{fujita_end--end_2019}
Y.~Fujita, N.~Kanda, S.~Horiguchi, Y.~Xue, K.~Nagamatsu, and S.~Watanabe,
  ``End-to-{End} {Neural} {Speaker} {Diarization} with {Self}-attention,''
  2019.

\bibitem{horiguchi_end--end_2020}
S.~Horiguchi, Y.~Fujita, S.~Watanabe, Y.~Xue, and K.~Nagamatsu, ``End-to-{End}
  {Speaker} {Diarization} for an {Unknown} {Number} of {Speakers} with
  {Encoder}-{Decoder} {Based} {Attractors},'' 2020.

\bibitem{horiguchi_towards_2021}
S.~Horiguchi, S.~Watanabe, P.~Garcia, Y.~Xue, Y.~Takashima, and Y.~Kawaguchi,
  ``\BIBforeignlanguage{en}{Towards {Neural} {Diarization} for {Unlimited}
  {Numbers} of {Speakers} {Using} {Global} and {Local} {Attractors}},'' 2021.

\bibitem{kinoshita_integrating_2021}
K.~Kinoshita, M.~Delcroix, and N.~Tawara, ``Integrating end-to-end neural and
  clustering-based diarization: {Getting} the best of both worlds,''
  \emph{Proc. ICASSP}, 2021.

\bibitem{kinoshita_advances_2021}
------, ``Advances in integration of end-to-end neural and clustering-based
  diarization for real conversational speech,'' \emph{Proc. Interspeech}, 2021.

\bibitem{yang_robust_2022}
C.~Yang and Y.~Wang, ``\BIBforeignlanguage{en}{Robust {End}-to-end {Speaker}
  {Diarization} with {Generic} {Neural} {Clustering}},'' in
  \emph{\BIBforeignlanguage{en}{Proc. Interspeech}}, 2022, pp. 1471--1475.

\bibitem{coria_overlap-aware_2021}
J.~M. Coria, H.~Bredin, S.~Ghannay, and S.~Rosset, ``{Overlap}-{Aware}
  {Low}-{Latency} {Online} {Speaker} {Diarization} {Based} {On}
  {End}-{To}-{End} {Local} {Segmentation},'' in \emph{Proc. ASRU}, 2021.

\bibitem{gruttadauria_online_2024}
E.~Gruttadauria, M.~Fontaine, and S.~Essid, ``\BIBforeignlanguage{en}{Online
  speaker diarization of meetings guided by speech separation},'' 2024.

\bibitem{fujita2_end--end_2019}
Y.~Fujita, N.~Kanda, S.~Horiguchi, K.~Nagamatsu, and S.~Watanabe,
  ``End-to-{End} {Neural} {Speaker} {Diarization} with {Permutation}-{Free}
  {Objectives},'' 2019.

\bibitem{xue_online_2021}
Y.~Xue, S.~Horiguchi, Y.~Fujita, S.~Watanabe, and K.~Nagamatsu, ``Online
  {End}-to-{End} {Neural} {Diarization} with {Speaker}-{Tracing} {Buffer},''
  2021.

\bibitem{xue_online_2021-1}
Y.~Xue, S.~Horiguchi, Y.~Fujita, Y.~Takashima, S.~Watanabe, P.~Garcia, and
  K.~Nagamatsu, ``\BIBforeignlanguage{en}{Online {Streaming} {End}-to-{End}
  {Neural} {Diarization} {Handling} {Overlapping} {Speech} and {Flexible}
  {Numbers} of {Speakers}},'' 2021.

\bibitem{horiguchi_online_2022}
S.~Horiguchi, S.~Watanabe, P.~Garcia, Y.~Takashima, and Y.~Kawaguchi, ``Online
  {Neural} {Diarization} of {Unlimited} {Numbers} of {Speakers},'' 2022.

\bibitem{han_bw-eda-eend_2021}
E.~Han, C.~Lee, and A.~Stolcke, ``\BIBforeignlanguage{en}{{BW}-{EDA}-{EEND}:
  streaming {END}-{TO}-{END} {Neural} {Speaker} {Diarization} for a {Variable}
  {Number} of {Speakers}},'' in \emph{\BIBforeignlanguage{en}{Proc. ICASSP}},
  2021, pp. 7193--7197.

\bibitem{zhang_fully_2019}
A.~Zhang, Q.~Wang, Z.~Zhu, J.~Paisley, and C.~Wang, ``Fully {Supervised}
  {Speaker} {Diarization},'' 2019.

\bibitem{schmidt_recurrent_2019}
R.~M. Schmidt, ``\BIBforeignlanguage{en}{Recurrent {Neural} {Networks}
  ({RNNs}): {A} gentle {Introduction} and {Overview}},'' 2019.

\bibitem{vaswani_attention_2023}
A.~Vaswani, N.~Shazeer, N.~Parmar, J.~Uszkoreit, L.~Jones, A.~N. Gomez,
  L.~Kaiser, and I.~Polosukhin, ``\BIBforeignlanguage{en}{Attention {Is} {All}
  {You} {Need}},'' 2017.

\bibitem{cho_learning_2014}
K.~Cho, B.~van Merrienboer, C.~Gulcehre, D.~Bahdanau, F.~Bougares, H.~Schwenk,
  and Y.~Bengio, ``\BIBforeignlanguage{en}{Learning {Phrase} {Representations}
  using {RNN} {Encoder}-{Decoder} for {Statistical} {Machine} {Translation}},''
  2014.

\bibitem{williams_learning_1989}
R.~J. Williams and D.~Zipser, ``A {Learning} {Algorithm} for {Continually}
  {Running} {Fully} {Recurrent} {Neural} {Networks},'' \emph{Neural
  Computation}, vol.~1, no.~2, pp. 270--280, 1989.

\bibitem{hochreiter1997long}
S.~Hochreiter and J.~Schmidhuber, ``Long short-term memory,'' \emph{Neural
  Computation}, 1997.

\bibitem{przybocki_mark_2000_2001}
{Przybocki, Mark} and {Martin, Alvin}, ``2000 {NIST} {Speaker} {Recognition}
  {Evaluation},'' 2001.

\bibitem{noauthor_talkbank_nodate}
\BIBentryALTinterwordspacing
``{TalkBank}.'' [Online]. Available: \url{https://talkbank.org/}
\BIBentrySTDinterwordspacing

\bibitem{landini_simulated_2022}
F.~Landini, A.~Lozano-Diez, M.~Diez, and L.~Burget,
  ``\BIBforeignlanguage{en}{From {Simulated} {Mixtures} to {Simulated}
  {Conversations} as {Training} {Data} for {End}-to-{End} {Neural}
  {Diarization}},'' 2022.

\bibitem{dai_transformer-xl_2019}
Z.~Dai, Z.~Yang, Y.~Yang, J.~Carbonell, Q.~V. Le, and R.~Salakhutdinov,
  ``\BIBforeignlanguage{en}{Transformer-{XL}: {Attentive} {Language} {Models}
  {Beyond} a {Fixed}-{Length} {Context}},'' 2019.

\end{thebibliography}

\end{document}